\documentclass[letterpaper, 10 pt, conference]{ieeeconf}  

\IEEEoverridecommandlockouts                              

\usepackage{graphicx} 
\usepackage{mathtools}

\usepackage{amsmath} 

\usepackage{textcomp}
\usepackage{tabularx,booktabs}
\usepackage{color}
\usepackage[pdftex,dvipsnames]{xcolor}
\usepackage{array}
\newcolumntype{P}[1]{>{\centering\arraybackslash}p{#1}}
\newcolumntype{C}{>{\centering\arraybackslash}X} 

\usepackage{tikz}

\newcommand*\annotatedFigureText[4]{\node[draw=none, anchor=south west, text=#2, inner sep=0, text width=#3\linewidth,font=\sffamily] at (#1){#4};}
\newenvironment {annotatedFigure}[1]{\centering\begin{tikzpicture}
\node[anchor=south west,inner sep=0] (image) at (0,0) { #1};\begin{scope}[x={(image.south east)},y={(image.north west)}]}{\end{scope}\end{tikzpicture}}
\newcommand*{\figtiTFlitefont}{\fontfamily{phv}\selectfont}

\title{\LARGE \bf
Merging Classification Predictions with Sequential Information for Lightweight Visual Place Recognition in Changing Environments
}

\author{Bruno Arcanjo$^{1}$, Bruno Ferrarini$^{1}$, Michael Milford$^{2}$, Klaus D. McDonald-Maier$^{1}$ and Shoaib Ehsan$^{1}$
\thanks{}
\thanks{$^{1}$B. Arcanjo, B. Ferrarini, K. D. McDonald-Maier and S. Ehsan are with the School of Computer Science and Electronic Engineering, University of Essex, United Kingdom {\tt\small (email: bq17319@essex.ac.uk; bferra@essex.ac.uk; kdm@essex.ac.uk; sehsan@essex.ac.uk)}}%
\thanks{$^{2}$M. Milford is with the School of Electrical Engineering and Computer Science, Queensland University of Technology, Brisbane, QLD 4000, Australia
        {\tt\small (email: michael.milford@qut.edu.au)}}%
}

\begin{document}

\maketitle
\thispagestyle{empty}
\pagestyle{empty}

\begin{abstract}
Low-overhead visual place recognition (VPR) is a highly active research topic. Mobile robotics applications often operate under low-end hardware, and even more hardware capable systems can still benefit from freeing up onboard system resources for other navigation tasks.
This work addresses lightweight VPR by proposing a novel system based on the combination of binary-weighted classifier networks with a one-dimensional convolutional network, dubbed merger. Recent work in fusing multiple VPR techniques has mainly focused on increasing VPR performance, with computational efficiency not being highly prioritized. In contrast, we design our technique prioritizing low inference times, taking inspiration from the machine learning literature where the efficient combination of classifiers is a heavily researched topic.
Our experiments show that the merger achieves inference times as low as 1 millisecond, being significantly faster than other well-established lightweight VPR techniques, while achieving comparable or superior VPR performance on several visual changes such as seasonal variations and viewpoint lateral shifts.

\end{abstract}

\section{Introduction}
\label{intro}
Visual place recognition (VPR) allows a system to localize itself in its operating environment using visual information, matching the currently observed place to a previously seen one. The localization information can then be used in downstream tasks such as Simultaneous Localization and Mapping (SLAM), allowing for internal map correction in mobile robotics systems \cite{ref:vpr-survey}.

VPR remains a difficult task with an array of non-trivial challenges. The same place can appear drastically different due to changes in illumination \cite{ref:illu_changes}, seasonal variations \cite{ref:season_changes}, dynamic agents \cite{ref:dyna_changes} and viewpoint variations \cite{ref:pov_changes}. In contrast, perceptual aliasing errors occur when two different places are identified as being the same due to visual similarities. Developing VPR techniques which are resilient to all or even just a subset of these errors is made harder when considering that many mobile robotic applications operate on hardware restricted platforms \cite{ref:res_hardware_1}, \cite{ref:res_hardware_2}, making computational efficiency an added important consideration. Due to size, payload limitation or cost, these platforms cannot carry powerful hardware such as graphic process units (GPUs) and hence rely on the development of efficient localization algorithms to autonomously navigate their environment. Moreover, developing lightweight techniques also benefits applications with more capable hardware, freeing up resources to be used in other important navigation tasks \cite{fischer2018icub, whereisyourplace}.

The different challenges that VPR presents have driven the development of several techniques, some being more resilient to certain visual variations than others \cite{ref:vpr_bench, whatmakesvprhard, unsu_comp_aware} and with different computational efficiencies. The availability of multiple VPR approaches with different characteristics has led to research in fusing different techniques into a single performing algorithm, enabling more accurate VPR \cite{ref:mpf}. However, most of the current fusion-based techniques \cite{ref:mpf, hier_mpf} heavily focus on VPR performance, with little attention given to computational efficiency, and might hence be unsuitable for resource constrained robotic platforms. 

This work aims to achieve highly efficient VPR, with low inference times. without relying on a dedicated graphics unit or any other dedicated computational hardware. We propose a novel lightweight system based on combining multiple compact classifiers using a one-dimensional convolutional neural network, dubbed  \textit{merger}. We start by introducing an efficient binary-weighted neural network \cite{ferrarini2020binary} as the baseline classifying unit. The merger network is then trained on score vectors obtained by multiple units, where the convolutional layer learns to relate the estimations of baseline models while incorporating sequential information. The proposed training schema relies on data augmentation to allow training both the baseline stage and the merger with a single environment sequence in an end-to-end fashion.

To the best of our knowledge, this is the first VPR technique to introduce the fusion of multiple baseline models with a learned network. We claim the following contributions in this manuscript: 

\begin{itemize}
    \item a lightweight, neural network with binary weights to be used as the baseline classifier of our VPR system.
    
    \item an one-dimensional convolutional neural network which efficiently combines the outputs of several baseline classifiers. 
    
    \item a data augmentation based training scheme for the proposed VPR system, allowing to train both the baseline classifiers and the merger network with a single environment traversal sequence.
\end{itemize}

The paper is structured as follows. In Section \ref{relatedwork} we give an overview of the state-of-the-art VPR techniques and prior work on combining multiple models into one system. In Section \ref{method} we detail our proposed system, from the binary-weighted classifier to the merger network and its training. Section \ref{exp_setup} details our implementation settings, the usage of datasets, and evaluation metrics. We present our results in Section \ref{results} and draw conclusions on this work and future directions in Section \ref{conclusions}.

\section{Related Work}
\label{relatedwork}
Several visual place recognition techniques have been proposed in the literature over the last decades. One popular approach is to compute local image descriptors from the most informative features in images from a training set \cite{ref:surf, ref:sift}, building an internal feature map. The map is then coupled with a retrieval algorithm, such as Bag-of-Words (BoW), which searches the feature map for the best match for an input image at runtime. While this pipeline has been heavily explored for VPR \cite{ref:fabmap, murillo2007surf}, it faces important challenges such as consistently selecting the most distinctive regions of an image and the long-term efficient storing and matching of the computed descriptors. The use of convolutional neural networks (CNNs) as feature extractors \cite{hou2015convolutional} addresses the first issue by having the network learn what information to extract from the image and techniques based on CNN feature extraction have recently achieved state-of-the-art VPR performance \cite{ref:netvlad, ref:hybridasmosnet, khaliq2018holistic}. However, the use of often large CNNs for feature extraction followed by searching the developed feature map for an optimal match makes these algorithms highly computationally intensive, and hence unsuitable for resource constrained robotic platforms. 

A possible alternative is to treat VPR as an image classification problem, where each place is treated as a class and each image is a training example belonging to a class. This approach has the advantage of encoding an image and inferring the place match in a single step, which can be designed to be highly efficient. Both DrosoNet \cite{voting_arcanjo} and FlyNet \cite{ref:flynetcann} are proposed classifier models designed for extremely lightweight VPR, with the downside of reduced performance. These small classifiers seem to present inconsistent performance, with two models of the same architecture and trained on the same environment sequence often outputting different place predictions for the same input image. \cite{voting_arcanjo} exploits this observation by using a large number of classifiers and combining their output with a handcrafted voting mechanism, trying to cancel out the weak spots of individual units.

Combining the estimations of multiple classifiers to improve classification performance has been heavily researched in the broader machine learning (ML) field \cite{ho1994decision, suen1992computer}, leading to powerful and popular ensemble algorithms \cite{chen2016xgboost}, \cite{randomforest}. For a combination of classifiers to perform well, it is important that their class estimations for a given input are, to some degree, different \cite{combining_classifiers}. Several approaches are possible to choose a set of baseline classifiers that meets this requirement. The most obvious of which is to simply use different classifier algorithms as baseline. If using multiple of the same baseline classifier, training the different models on different training sets will also generate variation in the predicted scores. When combining multiple identical neural networks, it is also possible to rely on the random weight initializations \cite{cho1995combining, hansen1990neural} to achieve different estimations, even when trained on the same training data. Despite of the chosen approach, it is also important that the baseline classifiers do not overfit the training data. Focusing on efficiency, using multiple fully binary neural networks has been shown to improve performance in image classification tasks \cite{zhu2019binary} while retaining the benefits of a low computational footprint and power usage. Binary neural networks seem particularly attractive to be combined in an ensemble, as each individual network is a weak but efficient classifier \cite{courbariaux2015binaryconnect}. The combination method itself has also been thoroughly explored in ML literature \cite{ho1994decision, roli2001methods}.The most common approach is to use a standard mathematical operation, such as multiplication or summation, on the outputs of the different classifiers. In contrast, it is also possible to train an output classifier using the scores of the baseline classifiers as inputs \cite{combiner_train_nottrain}. 

The combination of multiple baseline techniques to improve place matching performance has also been proposed for visual place recognition. \cite{ref:mpf} fuses the features obtained by multiple state-of-the-art (SOTA) VPR techniques in a Hidden Markov Model that also incorporates sequential information. Rather than parallel fusion, \cite{hier_mpf} proposes the use of an hierarchical usage of the baseline techniques. Measuring how different baseline techniques complement each other's weaknesses has also been a recent research topic \cite{ref:maria_compl}, with \cite{unsu_comp_aware} proposing a frame-by-frame selection of optimal techniques to combine. While fusion based techniques often achieve SOTA performance, they are demanding from a computational resource perspective, not being suitable for hardware constrained robotic applications. \cite{voting_arcanjo} shows that it is possible to design lightweight VPR systems based on merging classifiers as long as both the baseline unit and the combination method are efficient. Finally, to the best of our knowledge, all of the so far proposed fusing approaches for VPR are based on variations of the fixed mathematical operations discussed above, leaving the usage of a learned method for merging models for VPR largely unexplored.

In manuscript proposes using a learned classifier merging neural network for lightweight visual place recognition. We start from a compact baseline classifier with binary weights whose efficiency and compactness allow for the use of multiple units in parallel and whose outputs are then treated as inputs to a neural network, the merger. The merger network is based on a one-dimensional convolutional layer which combines the outputs of the baseline classifiers and relates the scores of nearby places in a single operation. We design an end-to-end training scheme based on data augmentation, allowing the baseline classifiers and merger network to be trained with a single environment traversal.

\section{Methodology}
\label{method}
The overall system consists of two main components: a binary-weighted neural network classifier used as the baseline unit and a convolutional neural network that merges multiple one-dimensional score vectors to achieve better classification performance. In this section, we start by explaining the two networks' architecture and then the merger networks' training process. The last two subsections cover the datasets that we utilize in our experiments and performance metrics used to evaluate our system against state-of-the-art techniques.

\subsection{Baseline Classifier}
\label{base_classifier}
\begin{figure}[thp]
\centering
\includegraphics[width=0.7\columnwidth]{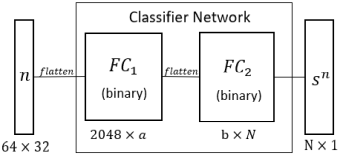}
\caption{Baseline classifier diagram.}
\label{fig:baseline_diagram}
\end{figure}
Our proposed baseline classifier is designed specifically to address lightweight VPR using the combination of multiple classifiers. Firstly, the baseline model must have low inference times to allow for efficient performance while using multiple models in parallel. The architecture of the baseline classifier can be seen in the diagram of Fig. \ref{fig:baseline_diagram}. The input to the model is an one dimensional vector containing all the pixels from an input image \(n\) which has been resized to 64 by 32 pixels in grayscale. The model comprises two fully-connected layers, \(FC1\) and \(FC2\), both with binary weights, allowing for less memory usage. The width \(a\) of the first dense layer \(FC1\) is a hyperparameter which controls the amount of neurons in the layer. The height \(b\) of \(FC2\) is the total amount of binary weights in \(FC1\) and is hence given by 
\begin{gather}
    b = 2048 \cdot a
\end{gather}
while its width corresponds to the total number of reference places \(N\), each considered a class. The output vector \(s^n\) contains the model's scores for each of the \(N\) places. The classifier achieves inference times as low as 1 millisecond, being on par with similarly lightweight VPR classifiers such as DrosoNet \cite{voting_arcanjo} and FlyNet \cite{ref:flynetcann}.

Secondly, we consider previous work in ML literature on combining classifiers into ensembles for improving performance. As established in Section \ref{relatedwork}, the combination of classifiers works best when the baseline models disagree on some of the classification tasks. We take several steps to achieve such behaviour. 

\subsubsection{Small Architecture}
Large models tend to overfit more compared to those with fewer parameters \cite{hawkins2004problem}, leading to more similar classifiers as they converge to more similar weights. The simple architecture of our proposed baseline model, while mainly for efficiency purposes, has the added benefit of reduced overfitting. Furthermore, binary weights usually overfit less to training data. The binarization method introduces noise when computing gradients, acting as a form of regularization and enables better generalization to unseen data \cite{hubara2017quantized}.

\subsubsection{Random Initializations}
As a neural network classifier with two fully-connected layers, our model depends on the random initialization of its weight parameters. Even with the same architecture and training set, different weight initializations may result in two different classifiers \cite{combiner_train_nottrain}.

\subsubsection{High Training Dropout}
As previously mentioned, it is important to prevent overfitting the models on the training set as this would generate less different classifiers. In order to reduce overfitting, we leverage a large dropout \cite{srivastava2014dropout} rate (\(>70\%\)) at training time, applied to the output of \(FC1\), as detailed in section \ref{model_settings}. It is also worth noting that the weights that get dropped are random on each training iteration, further helping with differentiating the baseline models.

\subsection{Convolutional Merger}
\label{conv_merger}

\begin{figure}[t]
\centering
\includegraphics[width=1\columnwidth]{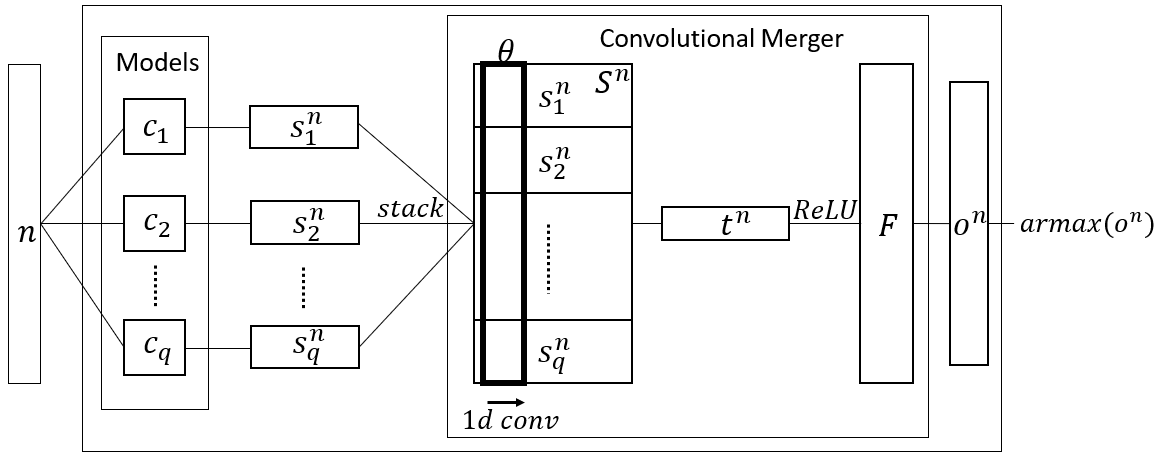}
\caption{Convolutional merger system.}
\label{fig:merger_diagram}
\end{figure}


As shown in the diagram of Fig. \ref{fig:merger_diagram}, the merger network consists of two layers: a one-dimensional convolution layer, followed by a \(ReLU\) activation, and a fully-connected layer. In the following paragraphs, we detail the prediction process for a currently observed place \(n\), with \(N\) total previously observed places.

The input to the network, matrix \(S^n\), is composed of vertically stacked score vectors \(s^n\) output by \(q\) classifiers when making a prediction on input image \(n\) and is of shape \(q \times N\).

The output vector of the convolution layer, \(t\), is of length \(l\), given by
\begin{gather}
\label{q_def}
    l = N + 1 - w
\end{gather}
where \(w\), a hyperparameter, is the width of the convolution.

Each element \(t_i\) of vector \(t\) is given by 
\begin{gather}
\label{conv_def}
    t_i = \sum_{a=0}^{q} \sum_{b=i}^{w+i} \theta_{ab} \cdot {S_{ab}}
\end{gather}
where \(a\) and \(b\) denote row and column numbers, respectively. The convolutional kernel weight matrix, \(\theta\), has shape \(q \times w\).

Equation \ref{conv_def} defines the one-dimensional convolution operation. The height of the kernel \(\theta\) is the same as the height of the input matrix \(S\) and corresponds to the number of baseline models, \(q\). As such, the kernel moves only in one dimension: horizontally, from left to right. 

Vector \(t\) is passed through a \(ReLU\) activation before being used as input for a dense layer with weight matrix \(F\) of dimensions \(l \times N\), giving each a final score for each place, vector \(o\). 
\begin{gather}
\label{fc_mult}
    o = ReLU(t) \cdot F
\end{gather}
Finally, the index of the highest score in \(o\) is taken to be the predicted place.
\begin{gather}
\label{argmax}
    predicted = argmax(o)
\end{gather}
The defined one-dimensional convolution combines the outputs of the baseline classifiers along the vertical dimension. Furthermore, for \(w > 1\), it also relates the scores of \(w\) adjacent places, exploiting the sequential nature of the input images. In a single convolution operation, the classifier's scores are not only combined but also infused with sequential information. The merging procedure becomes significantly more computationally efficient than other model fusing techniques, such as the voting system \cite{voting_arcanjo}, as proven later in table \ref{efficiency_results}.

\subsection{Training Schema}
\label{merger_train}

As explained in \ref{conv_merger}, the input to the merger network is a matrix constructed by stacking output score vectors of two or more base classifiers for a given input image. As mentioned in Section \ref{relatedwork}, combining classifiers is only interesting if they have, to some degree, different output estimations, with several possible approaches to generate different classifiers being highlighted in the ML literature. In our system, we use multiple of the same lightweight classifier network with some of these approaches built-in, detailed in section \ref{base_classifier}. Furthermore, we develop our system to be fully trained, baseline and merger networks, with a single training sequence. Collecting multiple traversals of an environment is often unpractical in real-world applications. The following paragraphs explain the process of collecting score vectors from the baseline classifiers suitable for training the merger network.

\subsubsection{Generating Score Vectors} 

\begin{figure}[th]
\centering
\includegraphics[width=0.98\columnwidth]{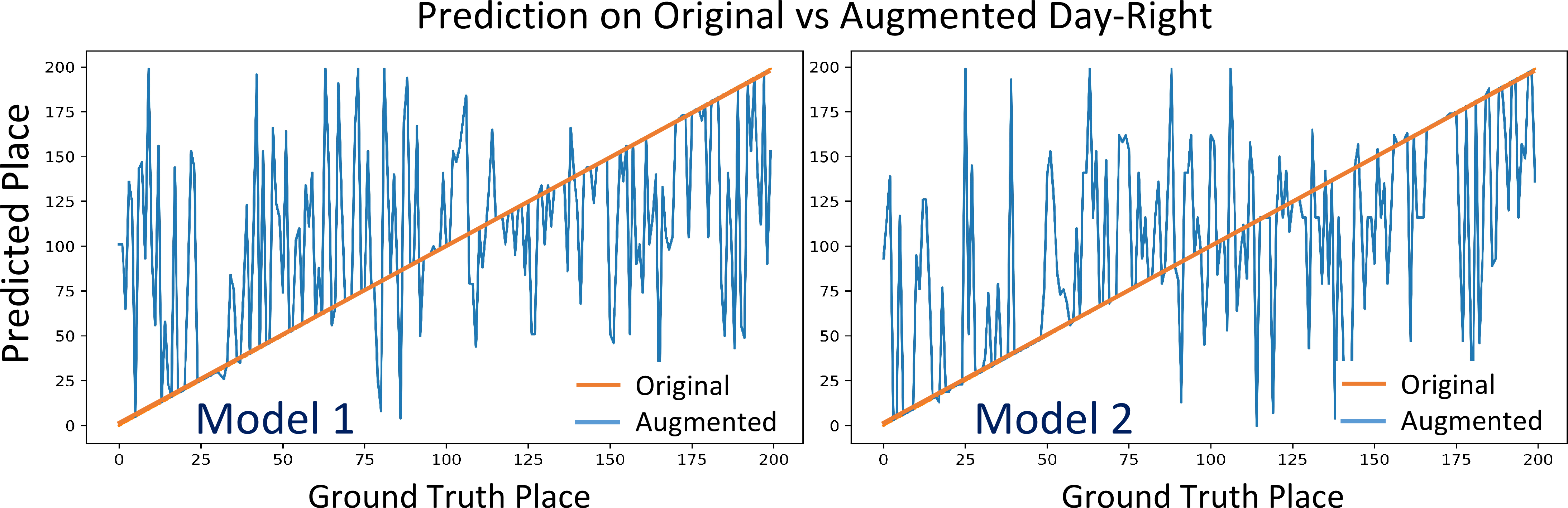}
\caption{Two model's predictions on original Day Right dataset used for training and the augmented version.}
\label{fig:predicted_simil}
\end{figure}
A set of score vectors can be obtained by inputting images into the baseline classifiers and collecting their estimations. For a set of score vectors to be suitable for training the merger network, there must be some disagreement on the predicted place between the different vectors. The orange lines In Fig. \ref{fig:predicted_simil} show that the trained classifiers achieve perfect predictions on the training set, returning identical outputs for each place in the training set. While this is expected behaviour, as the baseline models are able to converge during training, it makes these score vectors unusable for training the merger. Training the merger with such score vectors is like using only samples from the same class. Hence, training a neural network is not possible. To overcome this problem, we use data augmentation to inject some variability in the score vectors. First we generate a new training set of images by augmentation, as depicted in Fig 4. The newly generated subset is then used as input for the trained classifiers and the output score vectors are collected. The blue line in Fig. \ref{fig:predicted_simil} shows how the predictions made by two classifiers on the augmented set are not only imperfect but also different, with the miss classifications occurring on different frames.

\subsubsection{Data Augmentation Pipeline}
The data augmentation transformations are hyperparameters with the possibility to fine-tune the augmentation to replicate the expected visual changes in the environment. We choose a general approach of using the same two-step pipeline for every dataset, assuming no information about visual changes. Trivialaugment \cite{muller2021trivialaugment}, an efficient tuning-free algorithm, generates the first transformation, and the second transformation is a random horizontal flip of the image. 

\begin{figure}[th]
\vspace*{1ex}
\centering
\includegraphics[width=0.8\columnwidth]{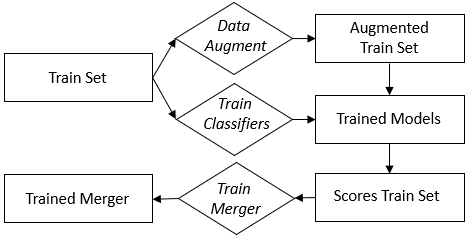}
\caption{Merger training diagram}
\label{fig:merger_training}
\end{figure}

\section{Experimental Setup}
\label{exp_setup}
We perform evaluation experiments with several state-of-the-art VPR techniques on multiple benchmark datasets. This section, explains our experimentation process by introducing the datasets, metrics and implementation settings used.

\subsection{Benchmark Datasets}
\label{datasets}
\subsubsection{Nordland Winter \& Fall}
The Nordland dataset \cite{ref:nord} includes four different seasonal train journeys. The dataset has been extensively used to assess a technique's resilience to appearance changes, with each traversal presenting various degrees of difficulty. In our experiments, we use the Summer sequence as the training set and evaluate performance on the Fall and Winter traversals, testing moderate and extreme appearance change resilience, respectively. We utilize 1000 images per traversal, with a matched place being considered correct if it corresponds to the exact ground-truth frame or immediately adjacent frames.

\subsubsection{Gardens Point}
The Gardens Point dataset was produced by handheld footage of the Queensland University of Technology's campus. We employ two of the traversals in the dataset to address strong, unilateral point-of-view variation with minimal appearance change. Both sequences are recorded during the day, with the testing sequence being translated to the right. The entirety of the 200 images per sequence are used, with a 2 frames error margin around the ground truth being allowed, as per previous research \cite{EP}.

\subsubsection{Oxford RobotCar}
Several traversals with different appearance challenges are available within the Oxford RobotCar dataset \cite{ref:robotoxford}. In particular, we use the dataset to evaluate resilience to illumination changes by using the CrossSeasons subset \cite{ref:cross_seasons}. The models are trained on 200 images collected at dusk and evaluated on 200 images of the same places collected during the day. As per previous research \cite{ref:prevres1}, the frame margin for error is 10 frames around the ground truth frame.

\subsection{Evaluation Metrics}
\label{metrics}
Visual place recognition is usually evaluated under an unbalanced dataset setting, with very few images belonging to the correct match class and the remaining being classified as incorrect. Precision-Recall (PR) curves are therefore suitable to evaluate VPR performance \cite{ref:pr_jus1, EP} and the area under these curves (AUC) is used to summarize the information given by these curves.

The efficiency of a technique is expressed with its computational time required to perform VPR and memory usage.

\subsection{Implementation Details}
\label{model_settings}
To select optimal hyperparameters for the architecture and training of our proposed system's components, we conduct a series of ablation studies in a grid-search fashion. In these ablation studies, we ran each configuration five times and collected the average performance as well as the standard deviation of the settings. Due to space constraints, we present the results of these tests for a single dataset. The same experiment was conducted on all datasets, and we make a general choice of settings for all datasets rather than fine tuning each one.

\begin{figure}[t]
\vspace*{1ex}
\centering
\includegraphics[width=0.48\columnwidth]{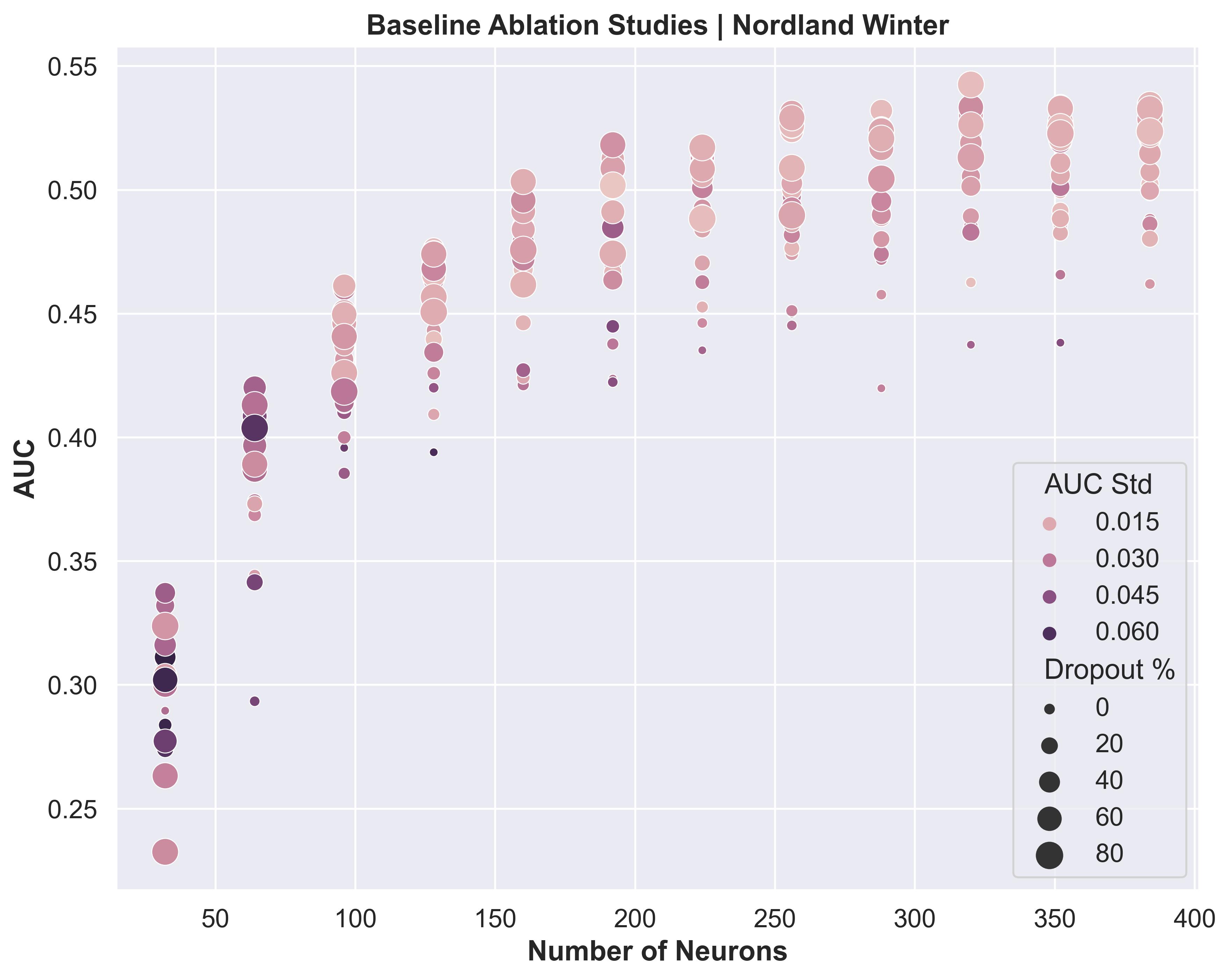}
\includegraphics[width=0.48\columnwidth]{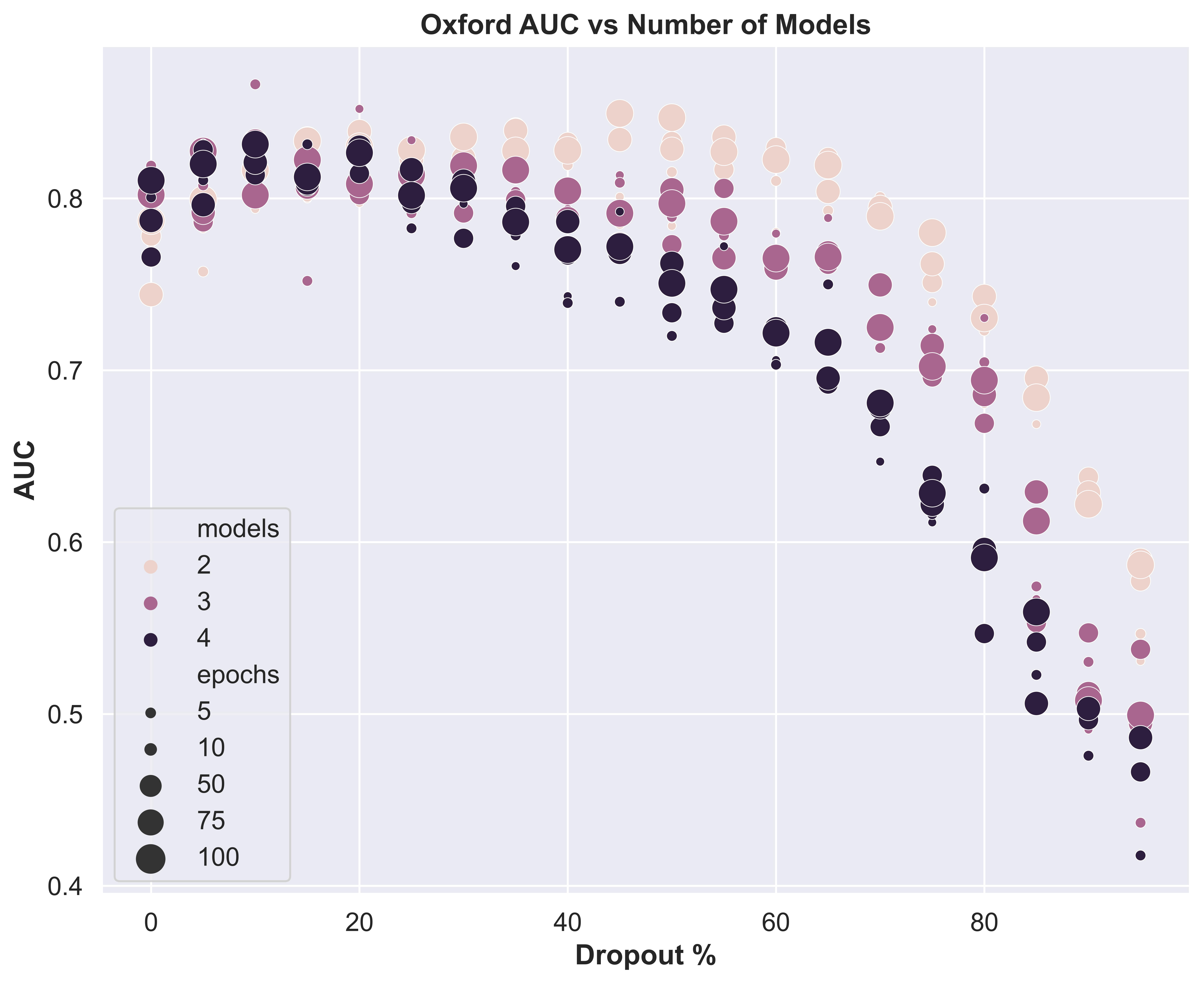}
\caption{Ablation studies to find optimal training parameters for the baseline classifier on the Winter dataset (left) and for the merger on the Oxford Robotcar dataset (right).}
\label{fig:baseline_merger_ablation}
\end{figure}

For the baseline classifier, we focus on selecting a suitable number of neurons and dropout rate at training time. In the left graph of Fig. \ref{fig:baseline_merger_ablation}, we observe that the performance starts out poor with a low number of neurons but quickly rises and stagnates at around 192 neurons, with minor improvements at higher numbers. Since the amount of neurons has a significant impact on the total number of weights in the classifier, see Section \ref{base_classifier}, we keep this parameter at 192. The training dropout percentage also seems to heavily influence the performance of the classifier. The performance peaks at around \(>70\%\) dropout ratio.

\begin{figure}[t]
\vspace*{1ex}
\centering
\includegraphics[width=0.68\columnwidth]{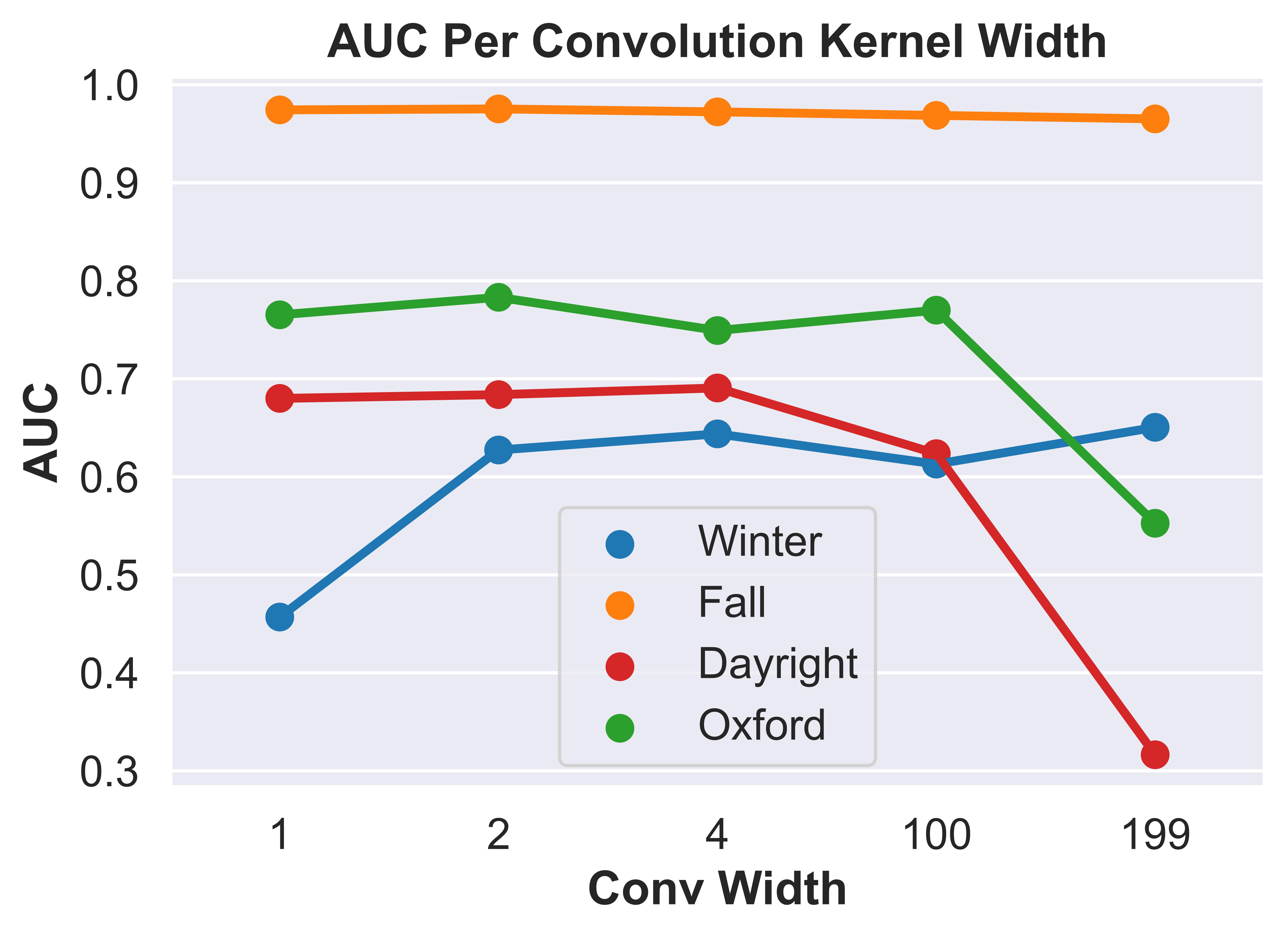}
\caption{Ablation study to find optimal values for the convolutional width kernel.}
\label{fig:width_ablation}
\end{figure}

Regarding the merger network settings, we also find reasonable choices for the training process settings, namely the number of training epochs and dropout rate percentage. In the right graph of Fig. \ref{fig:baseline_merger_ablation}, we observe that higher training epochs are required to achieve the best performance, and we choose to train our merger for 100 epochs. The merger's training dropout significantly impacts performance, with a value of around 30\% showing best results across all datasets. We observe also that 2 models achieve consistently better performance than 3 or 4 and we keep our system running on 2 baseline classifiers. Finally, we conduct a study to find optimal values for the width of the convolutional kernel \(w\), which dictates how many places are related by the convolution. Based on our results, shown in Fig. \ref{fig:width_ablation}, we choose a value of 4 for the kernel width.

For NetVLAD \cite{ref:netvlad}, ASMOSNet \cite{ref:hybridasmosnet}, CALC \cite{ref:calc}, and CoHOG \cite{ref:cohog}, we use the implementations provided in \cite{ref:vpr_bench}. For the Voting mechanism, we use the same setup as in \cite{voting_arcanjo}.


\section{Results and Discussion}
\label{results}
In this section we present the results achieved by our system in terms of visual place recognition ability and computational efficiency while comparing our results to other well-established lightweight VPR techniques.

\subsection{VPR Performance}
\label{performance_results}

\begin{figure*}[!ht] 
\vspace*{1ex}
\centering
\textsc{\small\figtiTFlitefont{PR curves comparison with lightweight VPR techniques}}\par\vspace*{3ex}
\begin{annotatedFigure}{\includegraphics[width=0.33\linewidth]{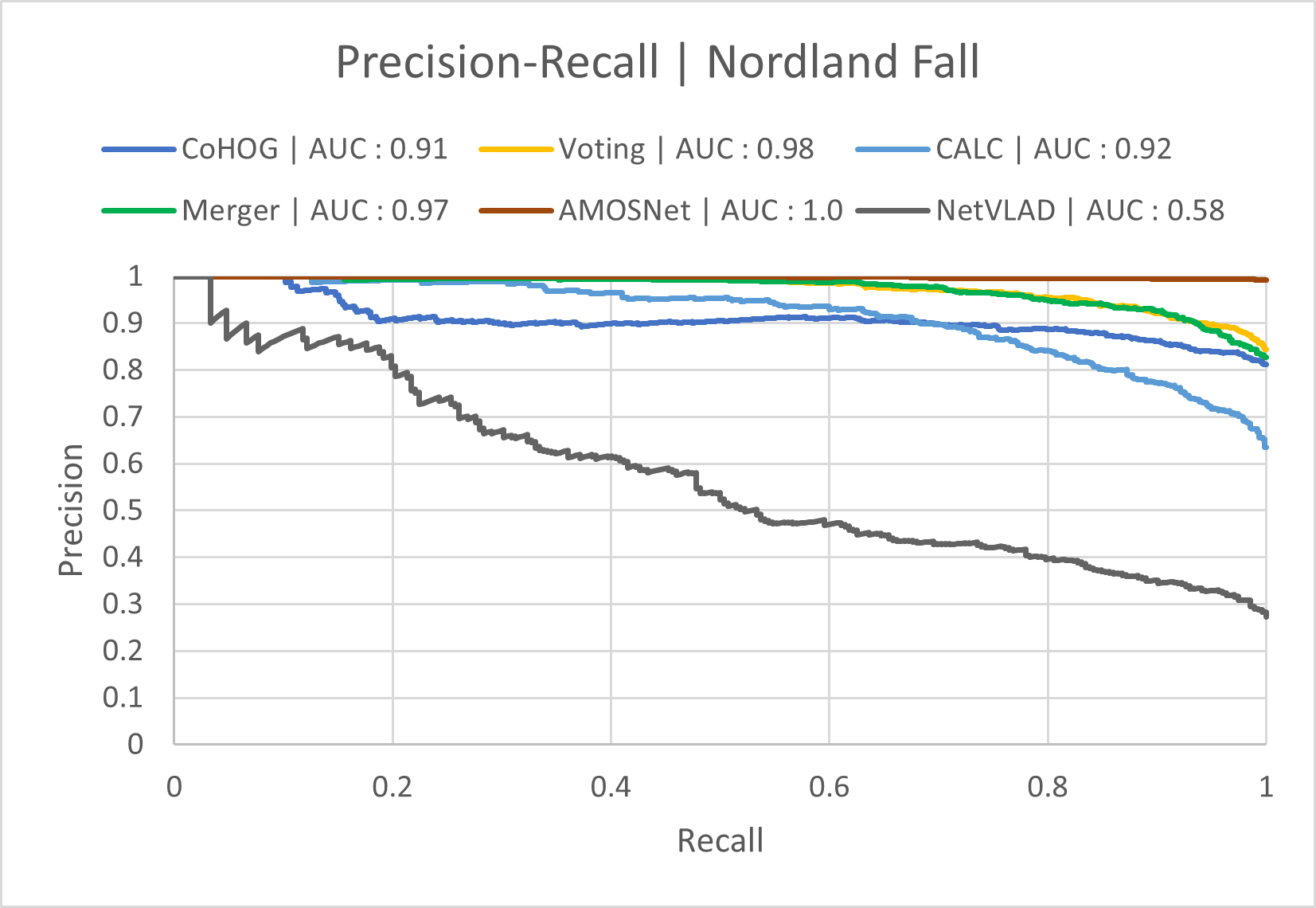}}
\annotatedFigureText{0.5,-0.07}{black}{0.00}{(a)}
\end{annotatedFigure}
\begin{annotatedFigure}{\includegraphics[width=0.33\linewidth]{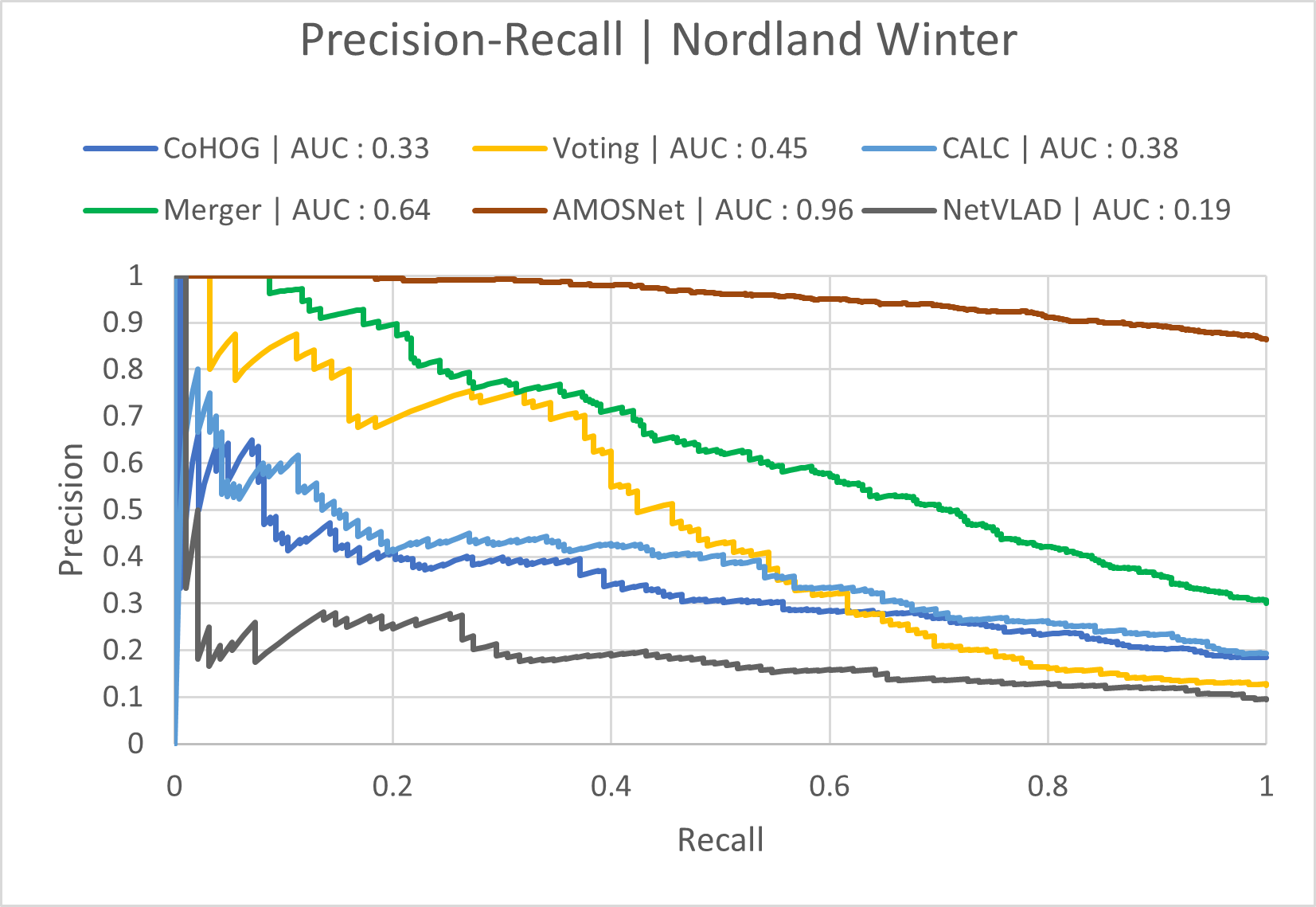}}
\annotatedFigureText{0.5,-0.07}{black}{0.00}{(b)}
\end{annotatedFigure}
\begin{annotatedFigure}{\includegraphics[width=0.33\linewidth]{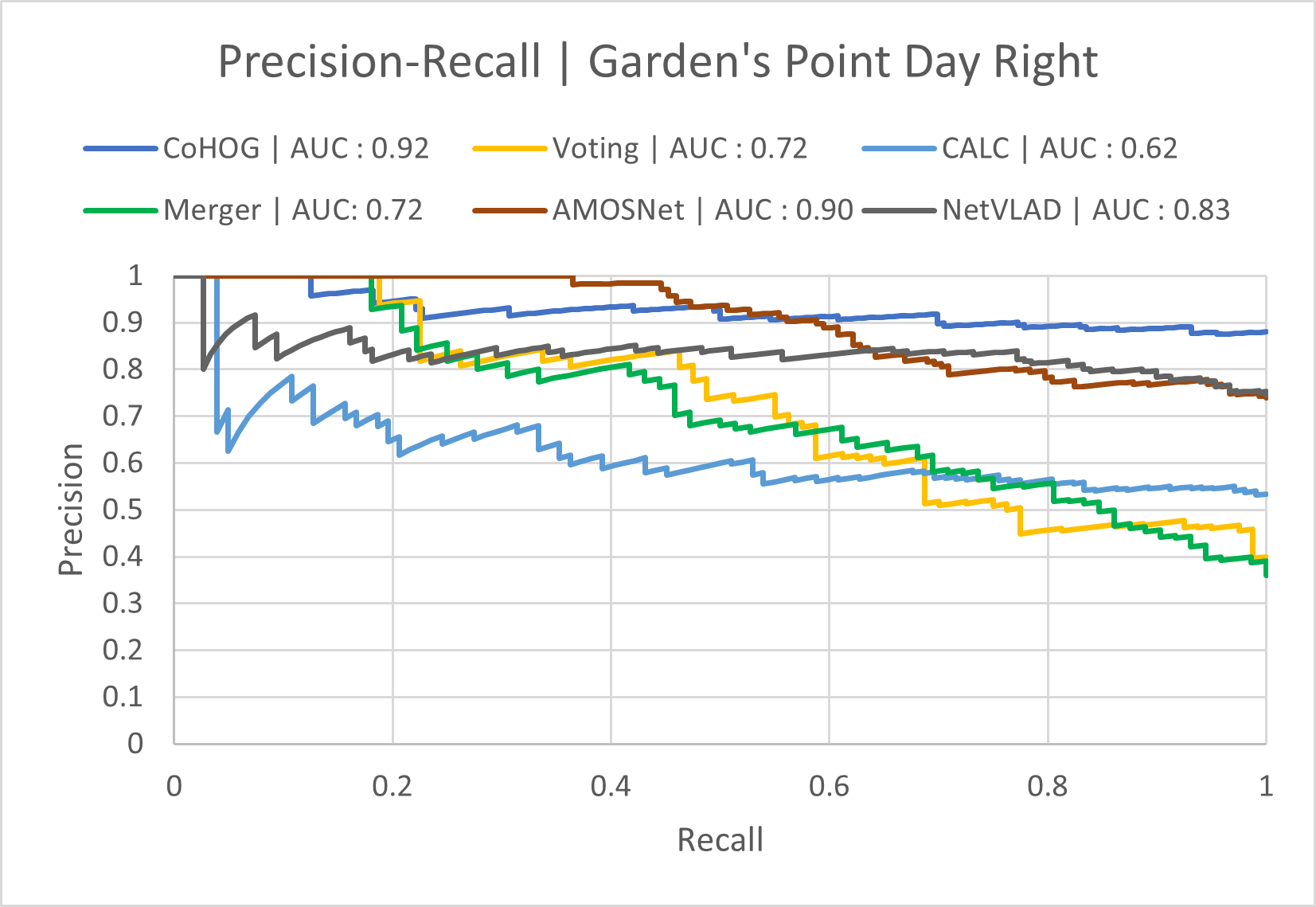}}
\annotatedFigureText{0.5,-0.07}{black}{0.00}{(c)}
\end{annotatedFigure} 
\begin{annotatedFigure}{\includegraphics[width=0.33\linewidth]{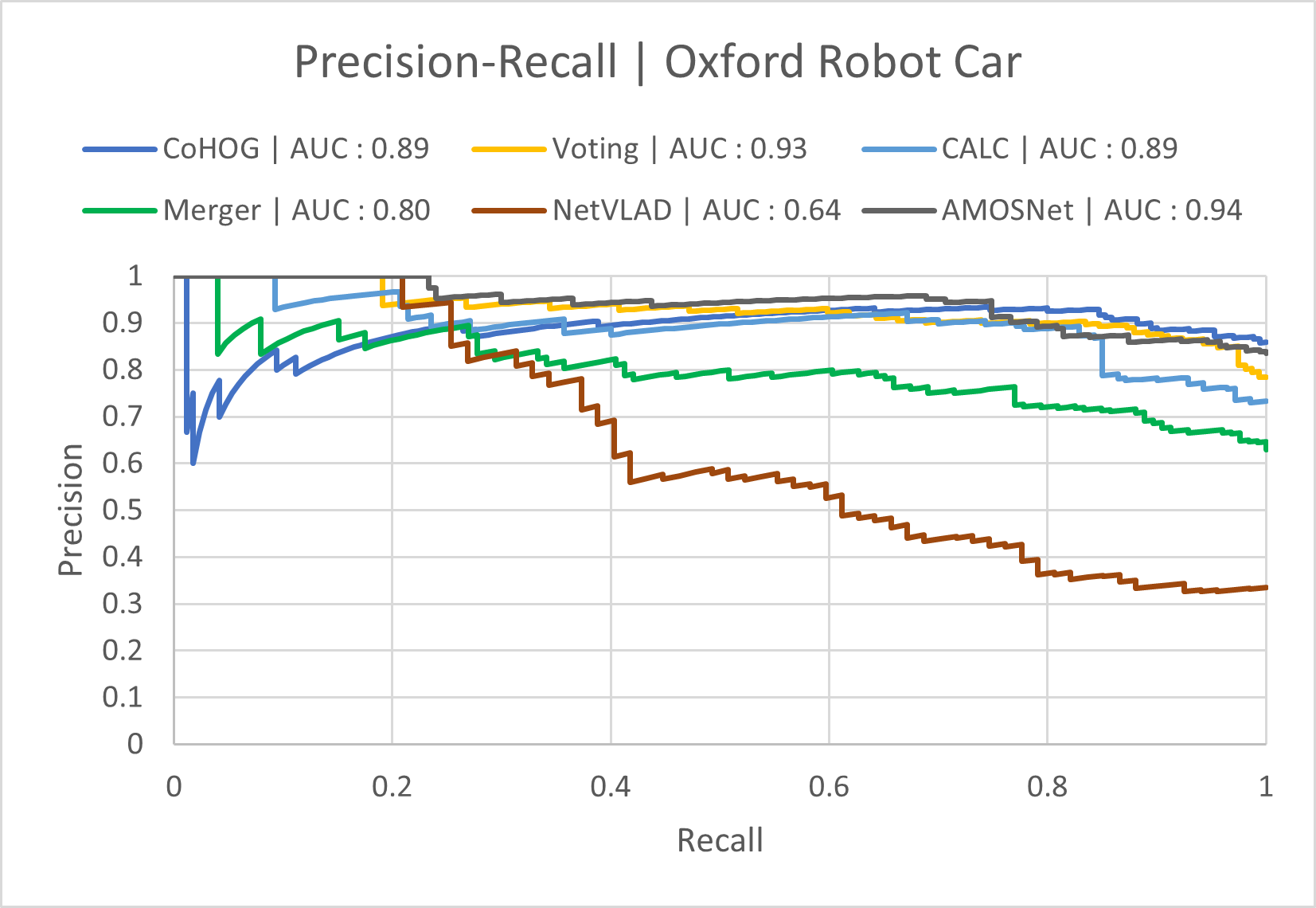}}
\annotatedFigureText{0.5,-0.07}{black}{0.00}{(d)}
\end{annotatedFigure}
\caption{Comparison of precision recall curves with respective AUCs for models which specifically target lightweight VPR.}
\label{fig:pr_curves}
\hspace{4ex} 
\end{figure*}

We compare the merger system's VPR performance against other state-of-the-art techniques designed for lightweight VPR, namely CoHOG, CALC and Voting. We also include  NetVLAD and AMOSNet in our testing, two non-lightweight VPR techniques, as to identify performance trade-offs.  The precision-recall curves for these models can be seen in the graphs of Fig. \ref{fig:pr_curves}.

Starting with the Nordland Fall dataset, which assesses moderate appearance changes, our technique achieves state-of-the-art performance, even when compared to the more computationally expensive techniques, observed in Fig. \ref{fig:pr_curves} (a).

When dealing with severe appearance changes, assessed by the Nordland Winter dataset, the merger achieves significantly better VPR performance than all the other lightweight techniques, only being outperformed by the demanding AMOSNet algorithm, as can be seen in Fig. \ref{fig:pr_curves} (b).

Regarding Garden's Point Day Right, which assesses viewpoint resilience with a lateral shift, the merger and voting achieve second highest AUC amongst the lightweight models, with CoHOG being the top performer, as can be seen in Fig. \ref{fig:pr_curves} (c). 

Finally, while the merger achieves a high AUC of 0.8 on the Oxford RobotCar dataset, it is outperformed by every model apart from NetVLAD, as is observable in Fig. \ref{fig:pr_curves}. Our model seems to struggle with the mixture of illumination changes with slight viewpoint variations.

\subsection{Computational Efficiency}
\label{efficiency_results}
\newcolumntype{M}[1]{>{\centering\arraybackslash}m{#1}}
\begin{table}[bp]
  \centering
  \caption{Inference Time and Model Size}
  \label{memory&time}%
    \begin{tabular}{M{1.5cm}M{1.5cm}M{1.5cm}M{1.5cm}}
    \toprule
Model     & Inference Time (ms) & FPS   & Size (MB) \\
\midrule
AMOSNet   & 886.09               & 1.13  & 61.44      \\
NetVLAD   & 741.53               & 1.35 & 16.38     \\
CoHOG     & 989.64               & 1.01  & 123.01     \\
CALC      & 56.93                & 17.57 & 4.26       \\
Voting    & 3.22                 & 310.56 & 6.19      \\
Merger    & 0.97                 & 1030.93 & 8.68      \\
\bottomrule
\end{tabular}
\end{table}%

We now discuss the results in Table \ref{memory&time} which summarizes the computational efficiency results, in terms of inference time and memory size, obtained by the merger and competing state-of-the-art techniques. These results were computed using the Winter dataset on an Intel I9 12900k CPU.

The merger has the lowest inference time of all the tested techniques. The next fastest technique is Voting, another lightweight combination based algorithm, and it is more than 3 times slower than our system while being outperformed with severe appearance changes and only achieving better VPR performance on the Oxford RobotCar dataset. CALC, a lightweight convolution neural network, only achieves outperforms our merger on the Oxford RobotCar while exhibiting a much higher inference time. CoHOG and NetVLAD show significantly longer inference times while being outperformed on the Nordland Fall and Winter datasets. AMOSNet is the top performer on all datasets with the exception of Garden's Point Day Right shift. However, it is three orders of magnitude slower than our proposed method, which might be preferred to AMOSNet to enable VPR on low-end hardware.

\section{Conclusions and Future Work}
In conclusion, we present a lightweight visual place recognition system based merging  binary-weighted neural network classifiers using a one-dimensional convolutional neural network. We design our models and training processes by considering previous research in combining classifiers from the machine learning field. Our technique achieves state-of-the-art performance amongst other lightweight VPR algorithms while retaining the fastest inference times among the tested techniques, making it an attractive option for hardware-constrained robotic platforms.

The use of a learned neural network to merge a number of baseline models is still largely unexplored in the visual place recognition field. Future work could focus on analysing the benefits of the merger network with other well-established VPR techniques, even if not specifically for lightweight VPR. 

\label{conclusions}





\bibliographystyle{IEEEtran}
\typeout{}
\bibliography{ref}

\end{document}